 \documentclass[smallabstract,smallcaptions]{dccpaper}

\usepackage{epsfig}
\usepackage{amsmath}
\usepackage{amssymb}
\usepackage{color}
\usepackage{url}

\usepackage{multirow}
\usepackage{graphicx}
\usepackage[export]{adjustbox} 
\usepackage{accsupp}
\usepackage{float}
\usepackage{eso-pic}

\newlength{\figurewidth}
\newlength{\smallfigurewidth}

\begin{document}

\AddToShipoutPictureBG*{%
  \AtPageLowerLeft{%
    \raisebox{1cm}{%
      \makebox[\paperwidth][c]{%
        \parbox{0.8\paperwidth}{\footnotesize\centering
          This is a post-peer-review, pre-copyedit version of an article
          published in DCC 2026. The final authenticated version
          is available online at \url{https://doi.org/10.1109/DCC66757.2026.00021}.
        }%
      }%
    }%
  }%
}

\title
{\large
\textbf{FSFVE: Few Shot Compressed Face Video Enhancement}
}

\author{%
Varun Ramesh Jois, Antonella DiLillo, and James Storer \\ 
{\small\begin{minipage}{\linewidth}\begin{center}
\begin{tabular}{c}
Brandeis University \\
Waltham, MA, USA \\
\{vjois,dilant,storer\}@brandeis.edu
\end{tabular}
\end{center}\end{minipage}}
}

\maketitle
\thispagestyle{empty}

\begin{abstract}
Videocalling has become a popular form of communication in the world today, with many companies providing free services for it. However, there are still millions of people around the world that experience poor quality videocalls due to limitations in bandwidth. This despite, most people having the required hardware. 

In this paper we present a novel framework for enhancing highly compressed videocalls. We show, that with as little as 10 frames of the face, we can rapidly (in under 100 seconds) train a model to enhance that instance of the videocall. The model can be trained either prior to or during the call, enhancing the rest of the call by producing better quality video. 
The video conferencing application need not be modified - it can be off the shelf with our system as a layer on top that trains quickly then simply lets the video conferencing application (e.g. Zoom) run as usual, where our system intercepts and improves images before they are displayed.
The model is designed to run in realtime on low-compute devices such as a typical laptop CPU. 
Experimentally, we show that the model significantly improves quality of compressed face video both quantitatively as well as perceptually. Code can be found at \url{https://github.com/varun-jois/FSFVE}.

\end{abstract}

\Section{Introduction}
Videocalling has increased dramatically in the past few years and is enabled by the near universal adoption of smartphones. However, there are large discrepancies when it comes to internet speeds. Millions of people around the world, particularly in the developing world, experience highly compressed videocalls with an associated diminished experience.

To improve received highly compressed video, there have been many solutions proposed in the computer vision literature such as denoising, deblocking, super-resolution etc. Many of these models do a great job of ameliorating compression artifacts, but almost all of these models have the same problem - they either do not run in realtime, a necessity for videocalls, or they require GPUs to run which is beyond the reach of the average consumer at the moment.

In this paper, we present a novel solution to the face video enhancement problem that runs in realtime at the receiving end on typical smartphones or laptops. Specifically, we investigate the question "Is it possible to build a small and fast model for an instance of a videocall?" If such a model exists, we could construct it either prior or during the call, enhancing the rest of the call producing better quality video. Our contributions are the following:
\begin{enumerate}
    \item We present our model \textbf{F}ew\textbf{S}hot Compressed \textbf{F}ace \textbf{V}ideo \textbf{E}nhancement (FSFVE); a model for realtime, low-compute (CPU) compressed face video enhancement that can be trained on as little as 10 frames of the subject.
    \item We show the effectiveness of our model, both quantitatively and qualitatively, at different high compression settings.
\end{enumerate}

\Section{Related Work}
Recent years have seen an explosion in research for the face restoration task. VQFR~\cite{vqfr} restores degraded facial images by replacing low quality encoded features with high quality vectors from a learned codebook. Its parallel decoder design fuses the codebook features and input information through a texture warping module employing the deformable convolution. GFP-GAN~\cite{gfpgan} uses generative facial priors (GFP) from a pretrained face GAN to restore details without relying on explicit geometric or reference priors. The model integrates these priors via channel-split spatial feature transform layers to produce the final output. DR2~\cite{dr2} diffuses input images into noisy images effectively nullifying any unknown degradation and then uses a trained diffusion model to make a coarse prediction. This coarse prediction is then improved upon using the VQFR model. The RTFVE~\cite{RTFVE} model performs face restoration for videocalls leveraging high-quality reference images; its dual branch design, allows reference features to be cached and its use of shuffle units~\cite{shufflenetv2} enable the model to perform in realtime on CPU. However, we show the RTFVE model underperforms in few-shot settings when limited training data is used which is the case for instance-based models. In MFQE~\cite{mfqe}, the model exploits quality fluctuations between frames in a video. It first detects Peak Quality Frames (PQFs) using a BiLSTM network and uses a multi-frame CNN to enhance lower quality frames with neighboring PQFs.

\Section{The FSFVE Framework} 
We want a model that performs well on the video producing significant improvement and also want the model to run in realtime (at least 24 frames/second) on CPU because if this is not met then the model becomes impractical. But strong models typically involve millions of parameters which require GPUs to run and typically do not run in realtime. However, we can exploit the fact that the frames of a videocall are highly correlated. By building a model for an instance of a call, we can constrain the problem to the nuances of the face at that moment, allowing us to build a strong model with fewer parameters. This model can then be used to enhance the subject for the duration of the call improving quality. This method is also practical for a multitude of reasons; quite often meetings are scheduled beforehand so the videoconferencing application can anticipate the need for an instance based model based on the duration of the call and the network connectivity. If enhancement is needed, the application can train the model either just before the call or at the beginning of the call when a few frames from the subject have been acquired. 

In summary, our model is based on these requirements:
\begin{enumerate}
    \item The model must run in realtime on CPU.
    \item The model must be able to learn patterns in the face with very little training data.
    \item The training must converge rapidly so that the model may be used for the remainder of the call.
\end{enumerate}
Convolutional Neural Networks (CNNs), the most popular type of model for vision tasks, generally do not meet these requirements. To give an example, a basic ResNet~\cite{resnet} with 5 convolution-relu-convolution blocks, a kernel size of 3, and 8 hidden channels runs in realtime on CPU but any model slightly larger becomes too slow. This model only contains around 6 thousand parameters and it has a very low ceiling when it comes to performance. Other changes can be made to the basic ResNet to make the model run faster, such as using depthwise separable convolutions, but as we shall see later, these models do not converge quickly breaking requirement 3.

The reason CNNs are popular for vision tasks is because of their ability to capture spatial relationships in the data. Another way to capture such relationships is by dividing the image into two dimensional DCT blocks. While not as effective as convolutions in capturing distant spatial relationships, working with DCT blocks nevertheless gives us the ability to model patches in the image. If, for instance, we use $8\times8$ blocks, each patch would contain $8^2*3=192$ numbers and we can construct a multilayer perceptron (MLP) to model the enhanced output of this patch. This type of model has a number of favorable properties - we get to capture some spatial information; we can construct a model with a large number of hidden layers or features giving us more parameters than a CNN and stronger modeling capabilities (requirement 2); inference is fast because we can take strides equal to the size of the DCT block (requirement 1). The final point is important because it is the main reason for our model's speed. When a CNN takes strides
greater than one the output spatial dimensions are reduced. 

To further improve modeling capabilities, we use the sine activation function~\cite{siren} to capture fine details in the face. This is because sine activations produce well behaved gradients that accurately capture high frequency details in the signal. Using the initialization scheme mentioned in \cite{siren}, preserves the distribution of activations throughout the network and ensures the input to each sine activation function is normally distributed with standard deviation 1.
This was found to yield fast and robust convergence which satisfies requirement 3. We can further improve convergence speed by making the model predict residuals instead of directly predicting the final output which can be loosely thought of as reversing the quantization step.

Mathematically, let the input frame be \(\mathbf{I} \in \mathbb{R}^{H \times W \times 3}\) with 3 color channels. 
We partition \(\mathbf{I}\) into \(N\) non-overlapping blocks of size \(B \times B\):
\[
\mathbf{I}_i^{(c)} \in \mathbb{R}^{B \times B}, \quad i = 1, \ldots, N, \quad c = 1, 2, 3
\]
For each block and channel, we apply the 2D DCT operator \(\mathcal{D}\):
\[
\mathbf{D}_i^{(c)} = \mathcal{D} \bigl(\mathbf{I}_i^{(c)}\bigr), \quad \mathbf{D}_i^{(c)} \in \mathbb{R}^{B \times B}
\]
We then vectorize each DCT block per channel and concatenate across the 3 channels into a single vector. This is the input to the MLP $\mathbf{x}_i^{0}$:
\[
\mathbf{x}_i^{0} =
\begin{bmatrix}
\text{vec}(\mathbf{D}_i^{(1)}) & 
\text{vec}(\mathbf{D}_i^{(2)}) & 
\text{vec}(\mathbf{D}_i^{(3)})
\end{bmatrix}
\in \mathbb{R}^{3B^2}
\]
Let us define the MLP model as a function \(\Phi: \mathbb{R}^{3B^2} \rightarrow \mathbb{R}^{3B^2}\).

The model \(\Phi\) consists of layers with sine activation:
\[
 \mathbf{x}^{(l)} = \sin\bigl(\mathbf{W}^{(l)} \mathbf{x}^{(l-1)} + \mathbf{b}^{(l)} \bigr)
\]

where \(\mathbf{W}^{(l)}\), \(\mathbf{b}^{(l)}\) are weights and biases at layer \(l\).

Each vectorized DCT block input passes through the MLP, producing residuals which get added to the input:
\[
\mathbf{y}_i = \Phi(\mathbf{x}_i^{0})+\mathbf{x}_i^{0}, \quad i=1,\ldots,N
\]
To produce the final enhanced patch, we reverse the concatenation, and vectorization steps and perform the 2D IDCT operation.

\Section{Implementation}
The FSFVE model can be trained either just before a videocall, or at the start
of a call. Only a few high quality images of the subject are required; e.g. 5 to 30 images making it few-short learning. Just before the call, or at the beginning of the call, a few seconds of high quality video of the user can be obtained; for instance 3 seconds, providing $3*24=72$ frames assuming a frame rate of 24 frames/second (FPS). This high quality video can quickly be re-encoded to a low quality video based on the settings for the videocall, and then with the low and high quality frames, the model is trained. 

The model may be trained on the sender's device and then transmitted to the receiver, or high quality frames can be sent to the receiver and the model trained there. A third option is for the sender to pass high quality images to a server, and then the server trains and passes the model to the receiver. The server can be used for cases where the sending and receiving devices are too slow for training (or even if not, to relieve them of the task). In any case, the size of the model is relatively small (about 3.5~MB) as are a few high quality frames, and once training is complete, the model can run in real-time on a typical laptop CPU.

Our system can be implemented as a layer on top of the videoconferencing application.
This layer can be created for any off-the-shelf videoconferencing application
such as Zoom. In fact, many applications such as Zoom and Google Meets provide APIs and SDKs to provide direct access to the frames.

\Section{Dataset}
All our experiments were performed using the publicly available DeepFakeDetection~\cite{DFDdataset} dataset which comprises of 363 videos of paid actors in 1080p quality. From these videos we chose the videos from the category \textit{"talking against wall"} because the subjects in these videos are seated at a table directly facing a camera closely mimicking a videocall. In total there were 27 videos averaging 972 frames each. While most frames in the dataset consist of frontal views, there are frames with partially-frontal views reflecting normal head movements that can be seen in a typical videocall.

To obtain low quality videos that replicate video quality under restricted bandwidth, we compressed our dataset with two widely used standards: H.264 and H.265, using the ffmpeg encoders libx264 and libx265. Compression levels were set to CRF values of 36, 40, and 44, which represent increasingly degraded quality. Since the CRF scale ranges from 0 (lossless) to 51 (worst quality), with the range of 17–28 usually considered acceptable, our choices produce videos that are similar in quality to those encountered in low bandwidth settings.

For all videos, we first took a $256\times256$ crop around the face from each frame by using the blazeface\cite{blaze} face detector. We then aligned the face so that the eyes were level and approximately at the same position for each frame. We did this by using a facial keypoint detetctor~\cite{face_recognition} to locate the eyes and then performing an affine warp. 

Since a crucial aspect of our model is to overfit on the videocall instance using a few frames, we had to decide on a method for extracting our training frames. Many selection criteria can be chosen such as random selection, selecting at fixed intervals etc. In order to get the most diversity in our training data, we used k-means clustering~\cite{kmeanspp}. After performing the face alignment, we passed each of these cropped and aligned frames to a face encoder~\cite{face_recognition} to get an encoding for each frame. With these face encodings, we performed k-means clustering with the number of cluster centers $k$ being the desired training size. We ran k-means 5 times in order to obtain the best cluster centers and the lowest inertia. Our training data was then formed by taking the frames (one from each cluster) that were closest to their respective cluster centers. For our experiments, we chose a training size of 30 for each video.

\begin{figure*}[tb]
    \centering
    \includegraphics[width=0.55\linewidth]{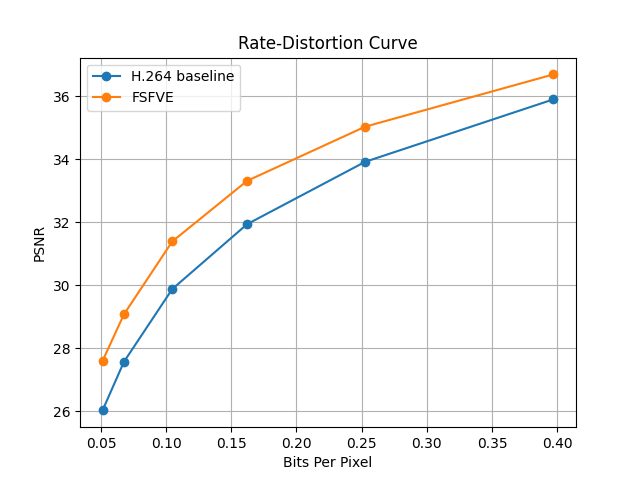}
    \EndAccSupp{}
    \caption{The Rate-Distortion Curve for our method.}
    \label{fig:rd}
\end{figure*}

\Section{Training}
For our method to be applicable in real-world scenarios, given that it is trained for a particular instance of a videocall, it needs to be trained quickly so that it may be used for the rest of the call. With these challenging constraints to consider, all our models were trained for only 100 epochs which took less than 100 seconds on average. We believe that 100 seconds of training time is reasonable since a large fraction of calls last for more than 10 minutes. To optimize our models, we used the RAdam optimizer~\cite{radam} with a learning rate of $10^{-4}$ and default parameters $\beta_1=0.9$, $\beta_2=0.999$. For our model's hyperparameters, we chose a block size of $8\times8$ and used 3 hidden layers, with each hidden layer having 512 features. 

We compared our model's output to the baseline compressed video, a basic ResNet \cite{resnet} and to the RTFVE model~\cite{RTFVE} which is the only other model we could find that runs in realtime on CPU. The ResNet hyperparameters were selected to enable real-time execution on a CPU. This configuration corresponded to a kernel size of 3, 5 residual blocks, and 8 hidden channels. Since the RTFVE model requires high quality reference images during inference, we modified the model to take no references while still having comparable speed and number of parameters as the original model. We call this RTFVE-0. For the ResNet and RTFVE-0 models the loss function that was used was the $L1$ loss. To ensure methodological fairness, all models, including competing methods, were trained using the same optimizer (RAdam) and under the same settings i.e. for each video in the dataset, an instance specific model was trained using 30 frames. 

\Section{Frequency Domain Focal Loss}
\label{sec:focal_loss}
Given our model's frequency domain training setup and constraints on training time, 
we accelerate convergence by reweighting the loss towards low frequency components, 
which have the greatest impact on perceptual quality. 
Formally, for each $8 \times 8$ block of DCT coefficients 
$\{c_{i,j}\}_{i,j=1}^{8}$, we define the weighted loss contribution as:

\[
\mathcal{L}_{\text{freq}} = \sum_{i=1}^{8}\sum_{j=1}^{8} w_{i,j} \, \ell(c_{i,j}, \hat{c}_{i,j}),
\]

where $\ell(\cdot,\cdot)$ is the base loss function (e.g., $L1$ or $L2$ loss) applied to the ground truth coefficient $c_{i,j}$ and its reconstruction $\hat{c}_{i,j}$. The weights are assigned according to:

\[
w_{i,j} = \frac{1}{Q_{i,j}},
\]

with $Q_{i,j}$ denoting the entry of the JPEG luminance quantization matrix~\cite{jpeg} 
at position $(i,j)$. Since $Q_{i,j}$ assigns smaller values to low frequency coefficients and larger values to high frequency ones, 
this reciprocal weighting naturally emphasizes the contribution of low frequency components, thereby prioritizing the image structures most critical to visual quality.
In our experiments we found the $L1$ loss to be the best base loss.

\begin{table}[th]
\caption{Performance scores with respect to video compression codec and compression rate. Higher the CRF, more the compression.}
\label{tab:scores}
\centering
\begin{tabular}{|c|c|c|c|c|c|}
\hline
\textbf{Codec} & \textbf{CRF} & \textbf{Model} & \textbf{PSNR (↑)} & \textbf{SSIM (↑)} & \textbf{LPIPS (↓)} \\
\hline\hline
\multirow{9}{*}{H.264} & \multirow{3}{*}{36} & H.264 (baseline) & 33.9231 & 0.9097 & 0.0979 \\
 & & ResNet & 33.9568 & 0.9134 & 0.1759 \\
 & & RTFVE-0 & 34.0147 & 0.9133 & 0.1401 \\
 & & FSFVE (Ours) & 35.0399 & 0.9257 & 0.1037 \\
\cline{2-6}
 & \multirow{3}{*}{40} & H.264 (baseline) & 31.9403 & 0.8837 & 0.1388 \\
 & & ResNet & 32.0716 & 0.8893 & 0.2301 \\
 & & RTFVE-0 & 32.1601 & 0.8887 & 0.1801 \\
 & & FSFVE (Ours) & 33.3174 & 0.9062 & 0.1225 \\
\cline{2-6}
 & \multirow{3}{*}{44} & H.264 (baseline) & 29.8786 & 0.8505 & 0.1935 \\
 & & ResNet & 30.1967 & 0.8618 & 0.2789 \\
 & & RTFVE-0 & 30.1563 & 0.8566 & 0.2328 \\
 & & FSFVE (Ours) & 31.3923 & 0.8771 & 0.1467 \\
\hline\hline
\multirow{9}{*}{H.265} & \multirow{3}{*}{36} & H.265 (baseline) & 33.5402 & 0.9068 & 0.1118 \\
 & & ResNet & 33.4028 & 0.9099 & 0.1962 \\
 & & RTFVE-0 & 33.6326 & 0.9116 & 0.1487 \\
 & & FSFVE (Ours) & 34.6758 & 0.9232 & 0.1112 \\
\cline{2-6}
 & \multirow{3}{*}{40} & H.265 (baseline) & 31.5015 & 0.8783 & 0.1552 \\
 & & ResNet & 31.6465 & 0.8869 & 0.2358 \\
 & & RTFVE-0 & 31.6737 & 0.8844 & 0.1911 \\
 & & FSFVE (Ours) & 32.8555 & 0.9011 & 0.1287 \\
\cline{2-6}
 & \multirow{3}{*}{44} & H.265 (baseline) & 29.4523 & 0.8444 & 0.2182 \\
 & & ResNet & 29.7738 & 0.8586 & 0.2835 \\
 & & RTFVE-0 & 29.6721 & 0.8517 & 0.2503 \\
 & & FSFVE (Ours) & 30.8846 & 0.8708 & 0.1559 \\
\hline
\end{tabular}
\end{table}

\Section{Quantitative Results}
To evaluate our models, we used both distortion and perceptual metrics. For distortion, we measured Peak Signal-to-Noise Ratio (PSNR) and Structural Similarity Index (SSIM)~\cite{ssim}, while for perceptual quality, we employed the Learned Perceptual Image Patch Similarity (LPIPS) metric~\cite{lpips}. Table~\ref{tab:scores} shows the scores for the models. For videos compressed with H.264 our model FSFVE, was able to improve PSNR upon the baseline by 1.11, 1.37 and 1.51 dB for CRFs 36, 40 and 44. Similarly, we find improvements for the SSIM metric. With respect to perceptual improvements, we see that our model is also able to do better on the LPIPS metric compared to the baseline for CRFs 40 and 44. We see similar improvements with the H.265 datasets suggesting the our method is agnostic to the video compression being performed. Figure~\ref{fig:rd} contains the rate-distortion curves of our model and the H.264 baseline. Our model's improvement over the baseline is more pronounced when the CRF is higher suggesting that our model does better under more compressed settings. The BD-Rate was found to be -25.1211~\% and the BD-PSNR was 1.3253~dB.

On the other hand, the ResNet and RTFVE-0 models struggle in these conditions barely outperforming the baseline. With the ResNet model, the dense convolution operation is slow on a CPU so we had to use a very small model of about 6 thousand parameters which was its shortcoming. For RTFVE-0, while it uses depthwise separable convolutions and has over 102 thousand parameters, the small training size and less training time made it difficult for the model to learn. Our model was able to outperform these models by over 1.1~dB with the same constraints showing the effectiveness of our model. A big reason for this was because our model is capable of fast inference (30.24 FPS) even with many more parameters - our model has slightly under 1 million parameters. 

Table~\ref{tab:train_size} shows the impact of training set size on the performance of unseen frames. This experiment was done for H.264, CRF 44 data. As expected, with more training data, we see better performance. With only 5 training images, our model is able to improve over the baseline by over 0.75~dB. And with 10 training frames, our model is able to improve the quality by over 1~dB showing the effectiveness of our method. 

To understand the performance improvement brought about by different aspects of our model design we performed an ablation study which can be seen in table~\ref{tab:ablation}. When working in the spatial domain, that is, the model takes $8\times8$ blocks of RGB values, the model is able to improve over the baseline by about 0.6 dB. By transforming the blocks into the frequency domain and giving equal weight to all the DCT coefficients, the performance further improves by about 0.6 dB. Finally, when transforming the blocks into the frequency domain and using the focal loss described earlier, performance improves by another 0.3 dB showing its effectiveness in speeding up model convergence. 

\begin{table}[tb]
\caption{PSNR with respect to number of training images. The PSNR of the baseline is 29.8786.}
\label{tab:train_size}
\centering
\begin{tabular}{|c|c|c|c|c|}
\hline
\textbf{Train Size} & \textbf{5} & \textbf{10} & \textbf{20} & \textbf{30} \\
\hline
FSFVE (Ours) & 30.6335 & 30.9452 & 31.0902 & 31.3923 \\
\hline
\end{tabular}
\end{table}

\begin{table}[tb]
\caption{Ablation study.}
\label{tab:ablation}
\centering
\begin{tabular}{|c|c|c|c|}
\hline
\textbf{Model} & \textbf{PSNR (↑)} & \textbf{SSIM (↑)} & \textbf{LPIPS (↓)} \\
\hline
H.264 (baseline) & 29.8786 & 0.8505 & 0.1935 \\
\hline
RGB spatial & 30.4675 & 0.8614 & 0.2358 \\
\hline
RGB freq. - L1 loss & 31.0783 & 0.8706 & 0.1589 \\
\hline
RGB freq. - Our loss & 31.3923 & 0.8771 & 0.1467 \\
\hline
\end{tabular}
\end{table}

\begin{figure*}[tb]
    \centering    
    \parbox[b]{0.15\linewidth}{\raggedright Low Quality Frames 1}\hfil
    \parbox[b]{0.15\linewidth}{\centering\includegraphics[width=\linewidth, valign=c]{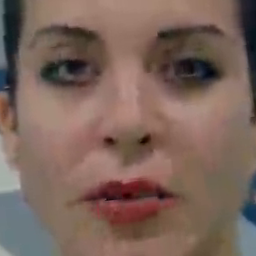}}\hfil
    \parbox[b]{0.15\linewidth}{\centering\includegraphics[width=\linewidth, valign=c]{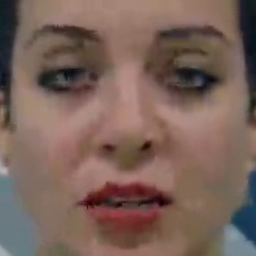}}\hfil
    \parbox[b]{0.15\linewidth}{\centering\includegraphics[width=\linewidth, valign=c]{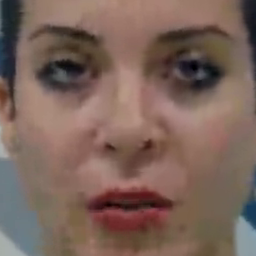}}\hfil
    \parbox[b]{0.15\linewidth}{\centering\includegraphics[width=\linewidth, valign=c]{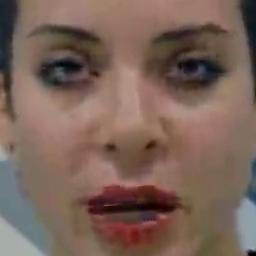}}\hfil
    \parbox[b]{0.15\linewidth}{\centering\includegraphics[width=\linewidth, valign=c]{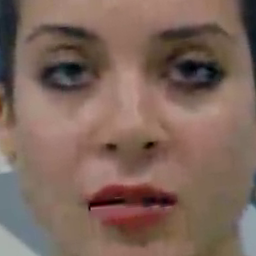}}\\[1ex]
    
    \parbox[b]{0.15\linewidth}{\raggedright Enhanced Frames 1}\hfil
    \parbox[b]{0.15\linewidth}{\centering\includegraphics[width=\linewidth, valign=c]{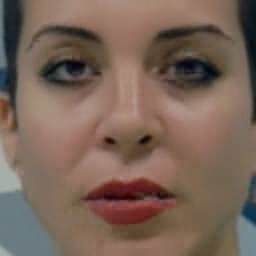}}\hfil
    \parbox[b]{0.15\linewidth}{\centering\includegraphics[width=\linewidth, valign=c]{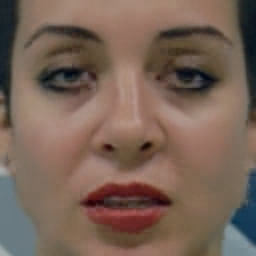}}\hfil
    \parbox[b]{0.15\linewidth}{\centering\includegraphics[width=\linewidth, valign=c]{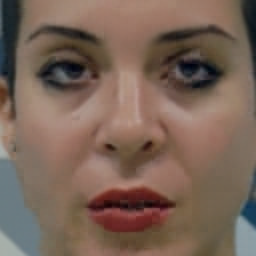}}\hfil
    \parbox[b]{0.15\linewidth}{\centering\includegraphics[width=\linewidth, valign=c]{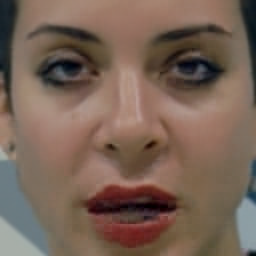}}\hfil
    \parbox[b]{0.15\linewidth}{\centering\includegraphics[width=\linewidth, valign=c]{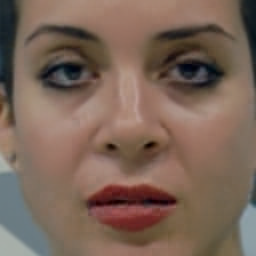}}\\[1ex]

    \parbox[b]{0.15\linewidth}{\raggedright Low Quality Frames 2}\hfil
    \parbox[b]{0.15\linewidth}{\centering\includegraphics[width=\linewidth, valign=c]{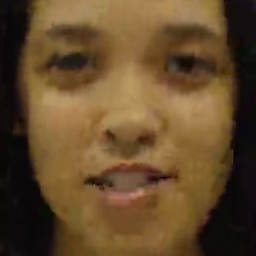}}\hfil
    \parbox[b]{0.15\linewidth}{\centering\includegraphics[width=\linewidth, valign=c]{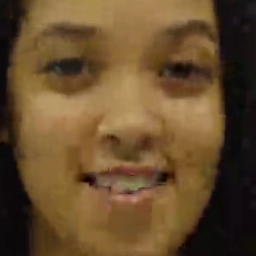}}\hfil
    \parbox[b]{0.15\linewidth}{\centering\includegraphics[width=\linewidth, valign=c]{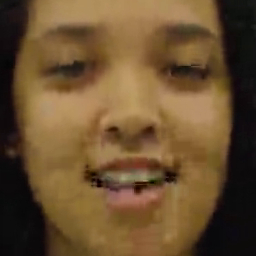}}\hfil
    \parbox[b]{0.15\linewidth}{\centering\includegraphics[width=\linewidth, valign=c]{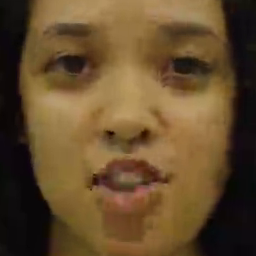}}\hfil
    \parbox[b]{0.15\linewidth}{\centering\includegraphics[width=\linewidth, valign=c]{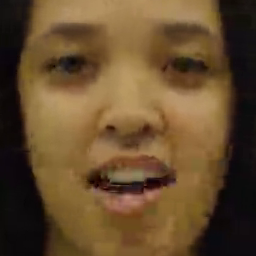}}\\[1ex]
    
    \parbox[b]{0.15\linewidth}{\raggedright Enhanced Frames 2}\hfil
    \parbox[b]{0.15\linewidth}{\centering\includegraphics[width=\linewidth, valign=c]{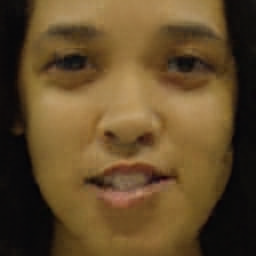}}\hfil
    \parbox[b]{0.15\linewidth}{\centering\includegraphics[width=\linewidth, valign=c]{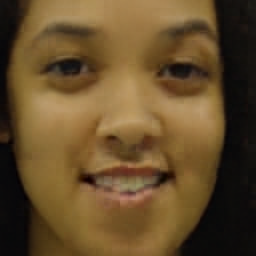}}\hfil
    \parbox[b]{0.15\linewidth}{\centering\includegraphics[width=\linewidth, valign=c]{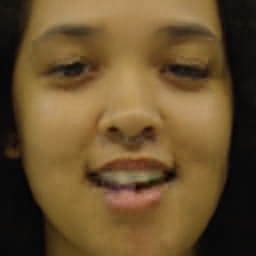}}\hfil
    \parbox[b]{0.15\linewidth}{\centering\includegraphics[width=\linewidth, valign=c]{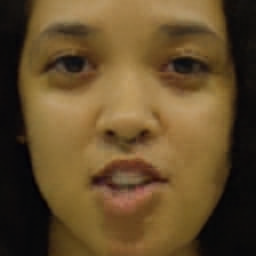}}\hfil
    \parbox[b]{0.15\linewidth}{\centering\includegraphics[width=\linewidth, valign=c]{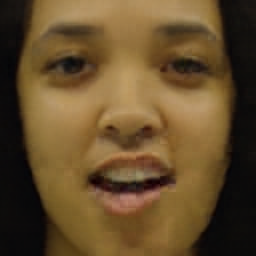}}\\[1ex]

    \EndAccSupp{}
    \caption{Stills from videos compressed at CRF 44.}
    \label{fig:crf44}
\end{figure*}

\Section{Qualitative Results}
Figure \ref{fig:crf44} shows stills from two videos compressed with H.264 at CRF 44. The figure also shows the output of our model for these compressed frames. The reader is encouraged to see this figure enlarged on a screen to notice the quality differences. In general, we see large compression artifacts in the low quality compressed frames around the eyes, nose and mouth of the subject. The skin is unnaturally textured and even the shapes of some facial components are distorted. 

In the first example, the eyes appear unclear with the black of the eye makeup blending with the eye color. There are clear signs of aliasing with the lips. The eyebrows lack definition and the skin appears distorted. 
In the second example, the skin is highly textured with speckle artifacts. There are blocking artifacts below the mouth that alter the shape of the chin. There also appear to be vertical lines.
Our model is able to remedy these artifacts generating a visually pleasing output with clearer skin and restores some of the fine details that were lost in the compression. Importantly, the enhanced output remains faithful to the identity in the video. The results are more pronounced on video and we encourage the reader to view our video demo at \url{https://sigport.org/documents/fsfve}.

\Section{Conclusion}
For millions of people around the world, videocall technology has not been fully appreciated due to limitations with internet speed. This is especially true in the developing world. In this paper, we addressed the problem of enhancing low quality videocalls, in realtime, on low-compute hardware such as laptop and mobile CPUs. 
We introduced a novel framework for training a model for an instance of a videocall where with only a few frames (10-30), we can train a model rapidly (under 100 seconds) and then use it to enhance the rest of the call. Experimentally, we verify the effectiveness of our model, showing a large improvement quantitatively and qualitatively over the baseline compressed video, as well as other models that run in realtime on CPU. 
\Section{References}
\bibliographystyle{IEEEtran}
\bibliography{refs}

\end{document}